\documentclass{article}
\usepackage{spconf,amsmath,graphicx}
\usepackage{multirow}
\usepackage{amsfonts}

\usepackage{graphicx}
\usepackage{subfigure}
\usepackage{caption}
\usepackage[utf]{kotex}
\usepackage{url}

\usepackage{xcolor}
\usepackage{pifont}
\newcommand{\cmark}{\ding{51}}%
\newcommand{\xmark}{\ding{55}}%

\newcommand{\ie}{\textit{i}.\textit{e}.}
\newcommand{\eg}{\textit{e}.\textit{g}.}

\title{Cut and Continuous Paste towards Real-time Deep Fall Detection}

\name{Sunhee Hwang, Minsong Ki, Seung-Hyun Lee, Sanghoon Park, and Byoung-Ki Jeon \thanks{This study was conducted using the airport abnormal behavior CCTV dataset constructed as part of the 「2019 Artificial Intelligence Identification and Tracking System Construction」 project of National IT Industry Promotion Agency (NIPA). Utilized data can be obtained from AI Hub (aihub.or.kr). }
}
\address{LG Uplus Corp. \\
\small \tt sunheehwang@lguplus.co.kr}

\begin{document}
%
\maketitle

\begin{abstract} 
Deep learning based fall detection is one of the crucial tasks for intelligent video surveillance systems, which aims to detect unintentional falls of humans and alarm dangerous situations.
In this work, we propose a simple and efficient framework to detect falls through a single and small-sized convolutional neural network.
To this end, we first introduce a new image synthesis method that represents human motion in a single frame.
This simplifies the fall detection task as an image classification task.
Besides, the proposed synthetic data generation method enables to generate a sufficient amount of training dataset, resulting in satisfactory performance even with the small model.
At the inference step, we also represent real human motion in a single image by estimating mean of input frames.
In the experiment, we conduct both qualitative and quantitative evaluations on \textit{URFD} and \textit{AIHub airport} datasets to show the effectiveness of our method.

\end{abstract}
\section{Introduction} 
\label{sec:intro}

Falling is one of the risky factors causing unintentional injury, particularly for the elderly, lone workers, and patients.
According to the WHO~\cite{who}, approximately 684,000 fatal falls occur, resulting in deaths each year globally.
To enable quick responses to such accidents, automated monitoring systems are getting widely used in nursing homes, manufacturing industries, and hospitals.
Specifically, a variety of fall detection methods are proposed to detect the human falling status based on diverse kinds of sensors, \eg, accelerometers, gyroscopes, biometric, and visual sensors.
Among these, a vision-based method is called an effective way, since it is non-invasive and there is no hassle of wearing devices. 
Moreover, it is easy to apply the method to existing video surveillance systems.

\begin{figure}[tb]

\begin{minipage}[b]{1.0\linewidth}
  \centering
  \centerline{\includegraphics[width=8.5cm]{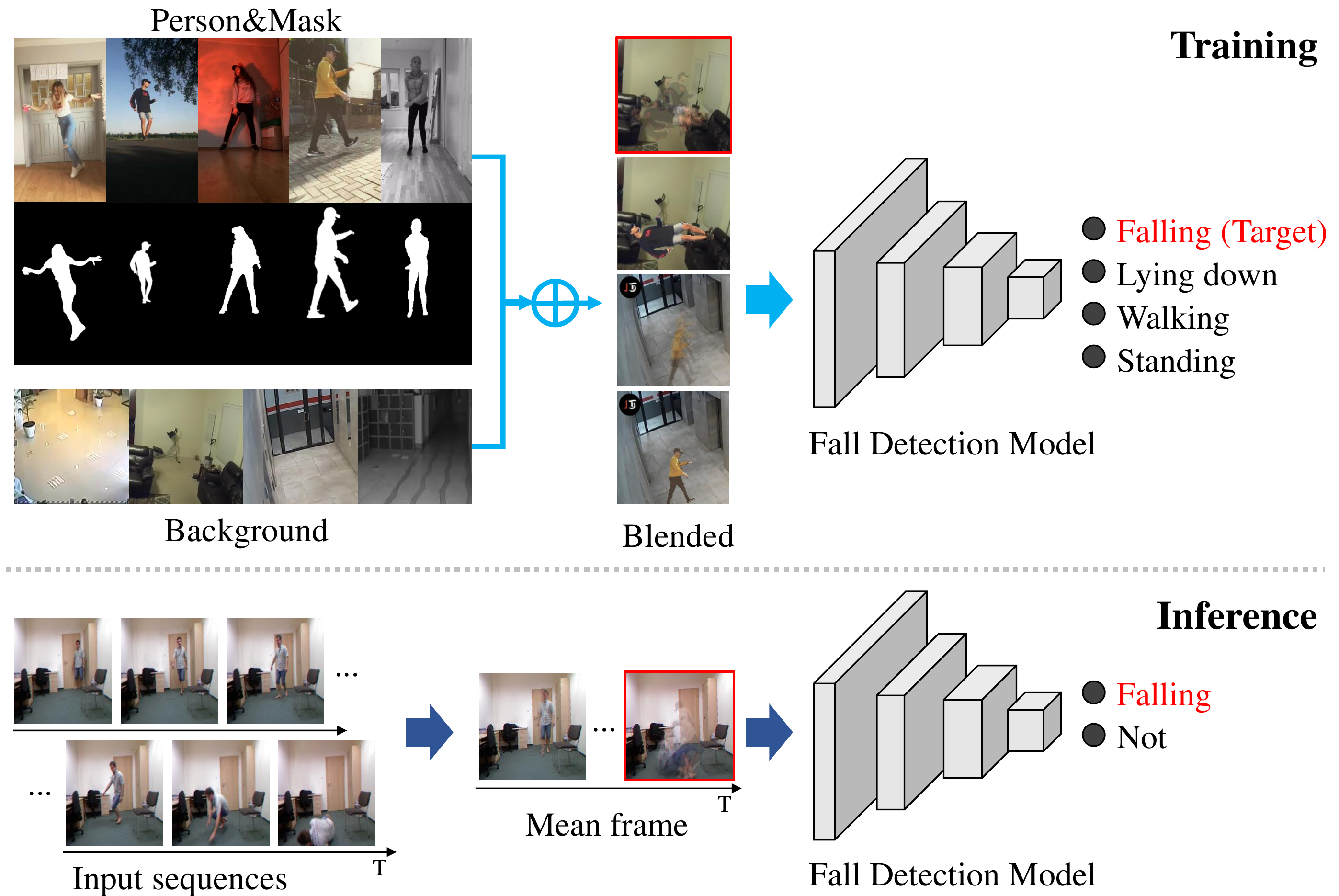}}
\end{minipage}
\caption{Overview of the proposed fall detection method. Synthetic images and real videos are used for training and test, respectively.}
\label{fig:overview}
\end{figure}

Most of the existing vision-based fall detection methods adopt a multiple-stage-based approach, which consists of two or more modules~\cite{pcnn,openpose_fall,maskRcnn,yolov3_deepsort,pipaf}.
Conventionally, the modules are composed as follows: person detection, tracking person, and falling state classification modules.
This approach can achieve high accuracy as it leverages the external knowledge of each module trained on large-scale datasets.
On the other hand, they have difficulty with system maintenance, since one module can affect the overall performance in terms of accuracy and inference time.
In addition, each module has to be updated to achieve better performance, which requires time-consuming data collection and labeling work.

\begin{figure*}[ht]
  \centering
  \subfigure[Person \& Mask]{%
    \includegraphics[width=0.151\textwidth]{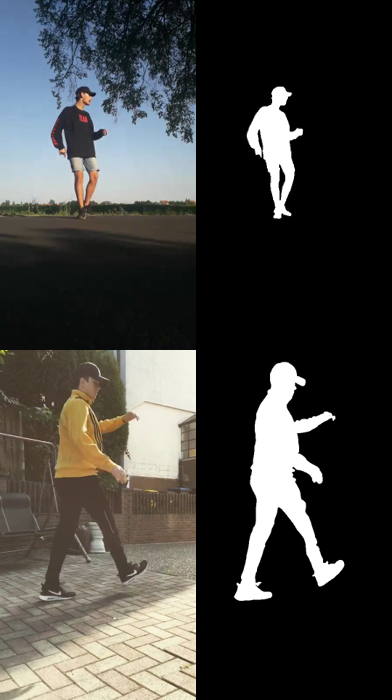}%
    \label{fig:img}%
  }%
  \subfigure[Background]{%
    \includegraphics[width=0.135\textwidth]{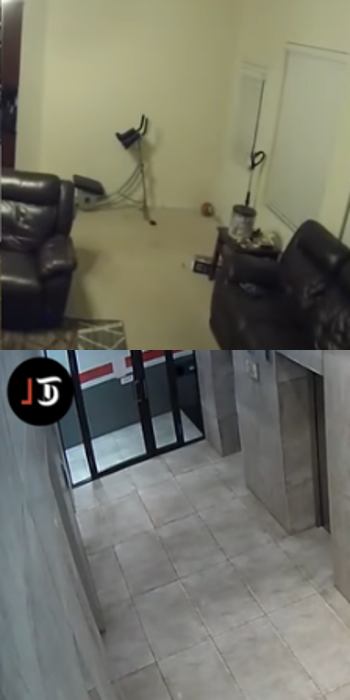}%
    \label{fig:bg}%
  }%
  \subfigure[$r_1$]{%
    \includegraphics[width=0.135\textwidth]{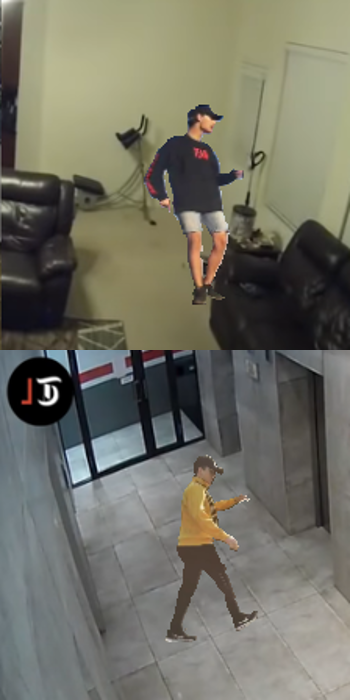}%
    \label{fig:r1}%
  }%
  \subfigure[$r_2$]{%
    \includegraphics[width=0.135\textwidth]{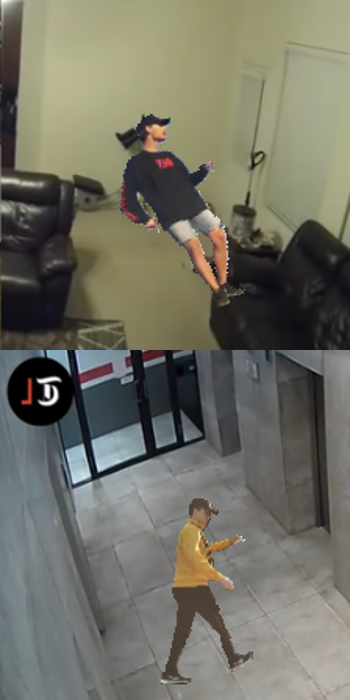}%
    \label{fig:r2}%
  }%
  \subfigure[$r_3$]{%
    \includegraphics[width=0.135\textwidth]{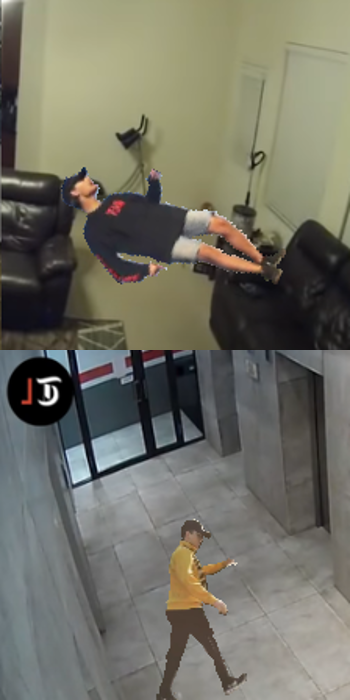}%
    \label{fig:r3}%
  }%
  \subfigure[$r_N$]{%
    \includegraphics[width=0.135\textwidth]{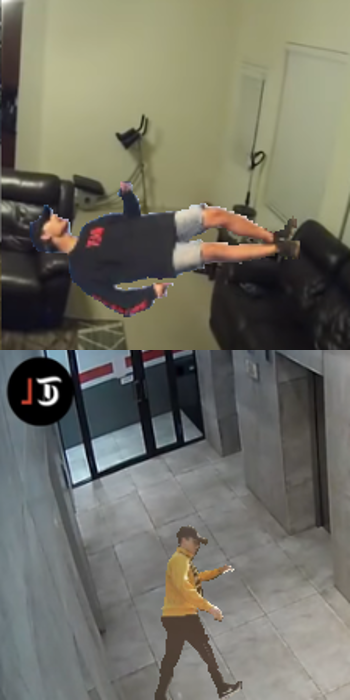}%
    \label{fig:r4}%
  }%
  \subfigure[$avg(\textbf{r})$]{%
    \includegraphics[width=0.135\textwidth]{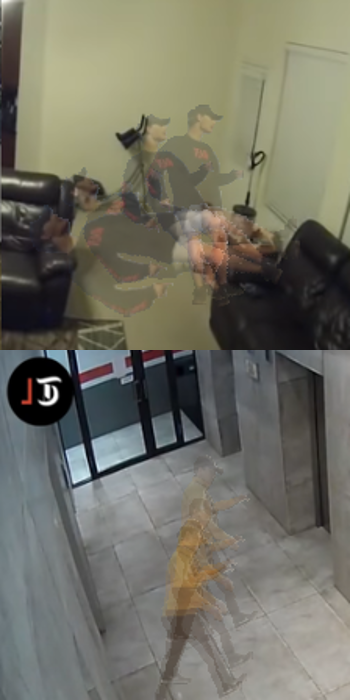}%
    \label{fig:blend}%
  }%
  \caption{The overall procedure of the proposed image blending method. Given person\&mask (a) and background image (b), we cutout person region from (a) and paste it into (b) after applying transform $\textbf{t}$: (c)-(f). Then, we average the images (c)-(f): (g). Finally, (f) and (g) are utilized to train the fall detection model as motionless and motional images, respectively.}
  \label{fig:imageblending}
\end{figure*}

Unlike the multiple-stage-based approach, several methods simplified the fall detection framework by adopting a single convolutional neural network (CNN) based models~\cite{mhi_vgg, bgs_lenet, vgg16_fall, flow_shallow}.
To this end, they conventionally train the fall detection models utilizing raw images or pre-processed motional images.
Then, they predict the falling state using the trained single CNNs.
This is a simple way but has the following difficulties.
Since most available fall detection datasets are obtained from limited person and place, they contain only a small number of samples for training the deep neural networks.
For that reason, it is not easy to obtain outstanding performance without utilizing external knowledge.
As an alternative way, several methods~\cite{mhi_vgg,vgg16_fall} adapt large CNNs such as VGG-16 to improve performance.
However, it is still hard to obtain excellent performance by training the deep models with small datasets.


From the existing methods, we observe that the fall detection models utilizing big-sized models or large-sized datasets tend to outperform others.
Unlike priors, we aim to design a new fall detection framework that performs well on the small-sized models without utilizing the large-sized datasets.
%
To this end, inspired by~\cite{cutpaste,copyandpaste}, we introduce an image synthesis method for generating large-sized training datasets.
We visualize the overall process in Figure~\ref{fig:overview}.
Specifically, given person and background images, we generate synthetic falling/non-falling samples.
To simplify the fall detection framework, we represent all the samples in a single frame.
Note that our method enables to generate the large-sized dataset without acting in dangerous situations for data acquisition.
Then, we train the single and small-sized models with the generated training data.
At the inference step, we also represent real motion in a single frame. 
For this, we estimate the mean of sequential images and predict its label through the trained model.
This approach simplifies the fall detection method as an image classification task and enables real-time processing. 

In experiment, we conduct two different fall detection datasets: \textit{URFD}~\cite{urfd} and \textit{AIHub airport}\footnote{https://aihub.or.kr/node/6258}.
To verify the superiority of our method, we compare the fall detection procedures with the existing methods.
We also compare the execution time and accuracy among various CNNs.
Overall, the proposed approach achieves best performance on EfficientNet-B0~\cite{efficientnet} with the accuracy of 97.14\% and 92.25\% on \textit{URFD} and \textit{AIHub} datasets, respectively.

In summary, our main contributions are as follows:
\begin{itemize}
    \item We propose a simple yet effective fall detection framework, reducing hardware and dataset requirements. 
    \item Through the quantitative and qualitative results on \textit{URFD} and \textit{AIHub} datasets, we validate the effectiveness of the proposed method. 
\end{itemize}

\section{Proposed Method} 
\label{sec:proposed method}

\subsection{Image Synthesis}
In this section, we describe the proposed image blending method, aiming to generate a large number of training samples.
As the main task of this study is fall detection, we first define four actions to distinguish between falling and non-falling.
The action categories are as follows: falling, walking, standing, and lying down.
Figure~\ref{fig:imageblending} shows the overall procedure of our blending method.
Given person images and their mask images (a), we cut them out and paste them into the background (b) to create the target images,~\ie, (f) and (g).

\begin{table*}[ht]
\centering
\caption{Fall detection performance comparison between our approach and exiting methods on the URFD dataset. The Second column denotes the architecture is composed of a single CNN or not. The third column indicates the usage of URFD dataset for model training, where $D_{tr}$ denotes the training dataset. We also note the number of parameters of single CNN-based models.}
\scalebox{0.86}{
\begin{tabular}{ccclccccc}
\hline
Ref.    & \multicolumn{1}{l}{Single CNN} & \multicolumn{1}{l}{URFD in $D_{tr}$} & \multicolumn{1}{c}{Architecture} & \multicolumn{1}{l}{Sensitivity} & \multicolumn{1}{l}{Specificity} & \multicolumn{1}{l}{Precision} & \multicolumn{1}{l}{Accuracy} & Params     \\ \hline
\cite{pcnn} & \xmark      & \cmark     & P-CNN + Tensorized-LSTM   & \textbf{100}              & 97.44      &  -   & \textbf{99.00}                  &    -   \\
\cite{openpose_fall}  & \xmark      & \cmark & OpenPose + LSTM    & \textbf{100}        & 96.40   &      -      & 98.20      &      -    \\
\cite{maskRcnn}  & \xmark     & \cmark     & Mask-RCNN + VGG16 + Bi-LSTM       & 91.80           & \textbf{100}      & \textbf{100}     & 96.70  & - \\
\cite{yolov3_deepsort}  & \xmark     & \cmark    & YOLO V3 + DeepSort + LSTM  & 93.10     &  -  & 94.80    &   -   &    - \\
\cite{pipaf} & \xmark     & \xmark   & PifPaf + Rule-based    & 83.33               & \textbf{100}      & -       &   -   &   -  \\
\hline
\cite{mhi_vgg} & \cmark      & \cmark     &  VGG16    &   -   &  -   &   -  & 92.34     & \multicolumn{1}{c}{138M} \\
\cite{bgs_lenet}  & \cmark      & \cmark     & BGS + LeNet     & 88.00     &      -   & 89.00       & 89.00             & \multicolumn{1}{c}{60K}  \\
\cite{vgg16_fall}  & \cmark & \cmark  & Modified VGG16  & \textbf{100}   & 92.00       &   -   & 95.00      & \multicolumn{1}{c}{138M+} \\
\cite{flow_shallow}  &\cmark  &\cmark   & Shallow-CNN    & 41.47           &    -   & 31.04       & 88.55         & \multicolumn{1}{c}{104K} \\
\textbf{Ours} & \cmark       & \xmark     & EfficientNet-B0      & 93.33           & \textbf{100}                    & \textbf{100}                  & \textbf{97.14}      & \multicolumn{1}{c}{4M}   \\ \hline
\end{tabular}
}
\label{tab:acc}
\end{table*}

Let a person image as $x$ and the corresponding binary mask as $m$. 
Before generating synthetic human motion images, we apply an online data augmentation approach with image transformation $f$ to $x$, where $f$ is composed of different kinds of randomly selected image jittering functions to ensure the diversity of the training images.
Afterwards, given $\hat{x}=f(x)$, we cutout the person segments using $m$, and then apply geometrical transformation $\textbf{t}\in\{t_1, t_2, ..., t_N\}$, where $\textbf{t}$ and $N$ denote a set of transformation functions to generate continuous human motion and the number of moments for the motion, respectively.
In other words, the transformation $\textbf{t}$ is designed to reflect the falling or walking motion, resulting in the $N$ sequential images.
Then, we individually paste the $N$ images, \ie, $t_1(\hat{x}), t_2(\hat{x}), ..., t_N(\hat{x})$, into background image $b$.
As a results, we attain the blended images $\textbf{r}\in\{r_1, r_2, ..., r_N\}$ representing a series of moment for human motion. 
While our goal is to train a single CNN without utilizing additional sequential modules, thus, we average $\textbf{r}$ (a set of $N$ blended images) to represent human motion in a single image.
Finally, we utilize $r_N\in\{lying~down, standing\}$ and $avg(\textbf{r})\in\{falling, walking\}$ for training samples of motionless image and motional image, respectively.

\subsection{Falling State Classification}
We then train the fall detection model on the generated synthetic dataset to classify the four categories (\ie, lying down, standing, falling, and walking).
Note that we utilize one of the small and single CNNs for fall detection.
Let $s^i\in\{r_N, avg(\textbf{r})\}$ is one of the randomly generated human motional image.
We compose the mini-batch with $M$ images $\textbf{s} \in \{s^1, s^2, ..., s^M\}$. 
Then, we train the model $g$ to predict corresponding human motional label $\textbf{y} \in \{y^1, y^2, ..., y^M\}$, where $g$ is optimized by the following cross-entropy loss function:

\begin{equation}
\mathcal{L}_{ce}= -\sum_{i=1}^{M}y^{i}log(g(s^i)).
\label{eq:s}
\end{equation}

For the inference, given real images from a camera, \ie, $\textbf{u} \in \{u^1, u^2, ..., u^\infty\}$, we start detecting falls from time $T$ using $N$ selected frames with a stride $k$. 
For example, given $\hat{\textbf{u}} \in \{u^{T-k\times N}, u^{T-k\times(N-1)}, ..., u^{T}\}$, we estimate the average of those frames to represent the actual motion sequence as a single frame.
Afterwards, we predict the target of human status using $g$ from $avg(\hat{\textbf{u}})$.

\section{Experiments} 
\subsection{Experimental Settings}
\textbf{Image blending:}
For the image blending, we utilize two datasets: \textit{HDD} \footnote{https://www.kaggle.com/constantinwerner/human-detection-dataset} and \textit{TikTok dancing} \footnote{https://www.kaggle.com/tapakah68/segmentation-full-body-tiktok-dancing-dataset} from Kaggle.
\textit{HDD} consists of 236 different scenes without humans. For this, we manually select ten background images in which no object is placed in the center of it.
\textit{TikTok dancing} consists of 2,615 person images with corresponding binary masks. 
For this, all images are utilized to generate a training set.
By blending images from both datasets with random image transformation, we obtain a tremendous amount of images for training.

\noindent\textbf{Models:}
To validate our method on the various kinds of single and low-cost CNNs, we used four ImageNet pre-trained models: ResNet-18~\cite{resnet}, MnasNet~\cite{mnas}, MobileNetV3~\cite{mobilenetv3}, EfficientNet18-B0~\cite{efficientnet}. 
For all models, we modify the last fully connected layers into the proper dimensions. 

\begin{figure*}
  \centering
  \subfigure[Falling]{%
    \includegraphics[width=0.43\textwidth]{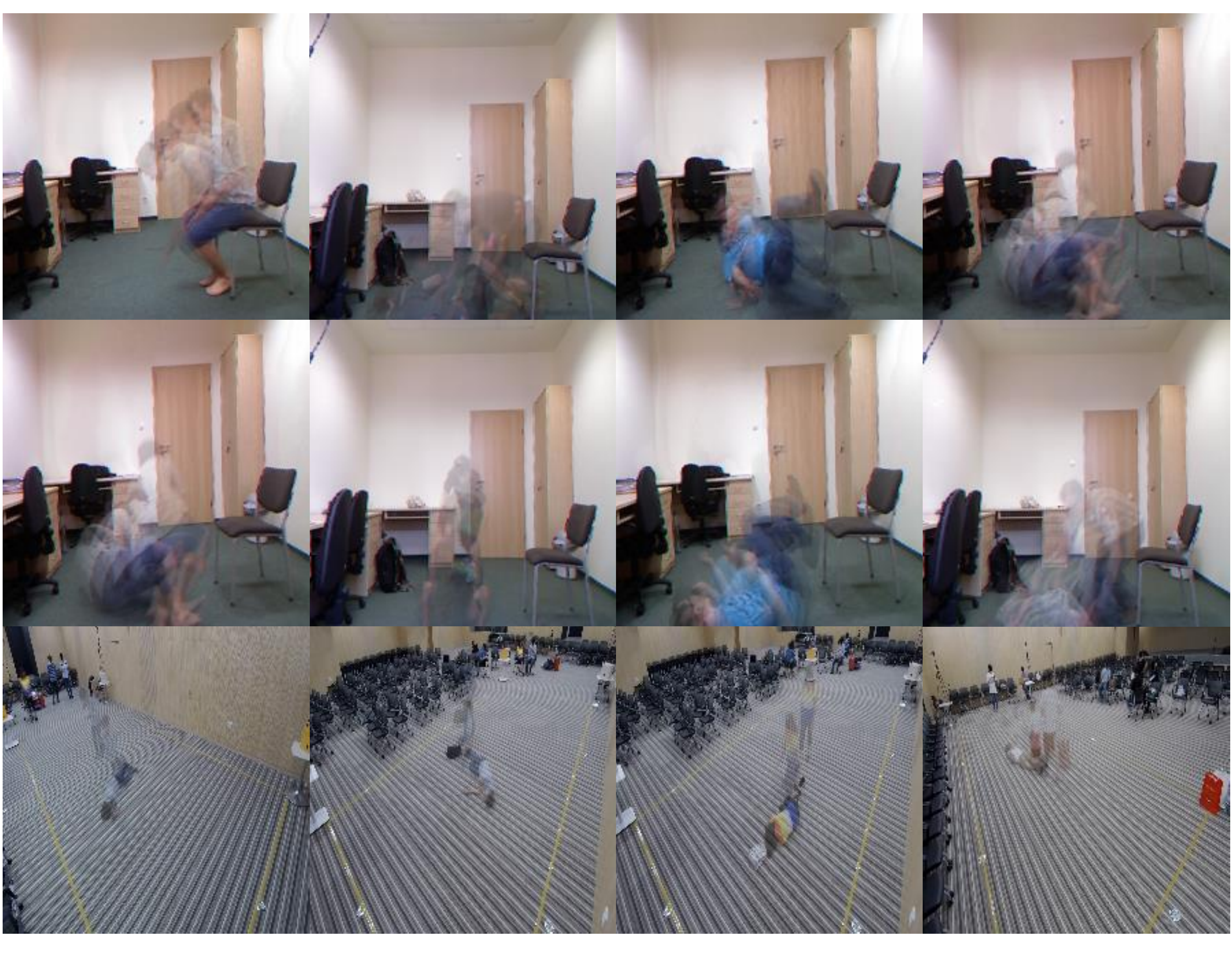}%
    \label{fig:out_falling}%
  }%
  \subfigure[Non-falling]{%
    \includegraphics[width=0.544\textwidth]{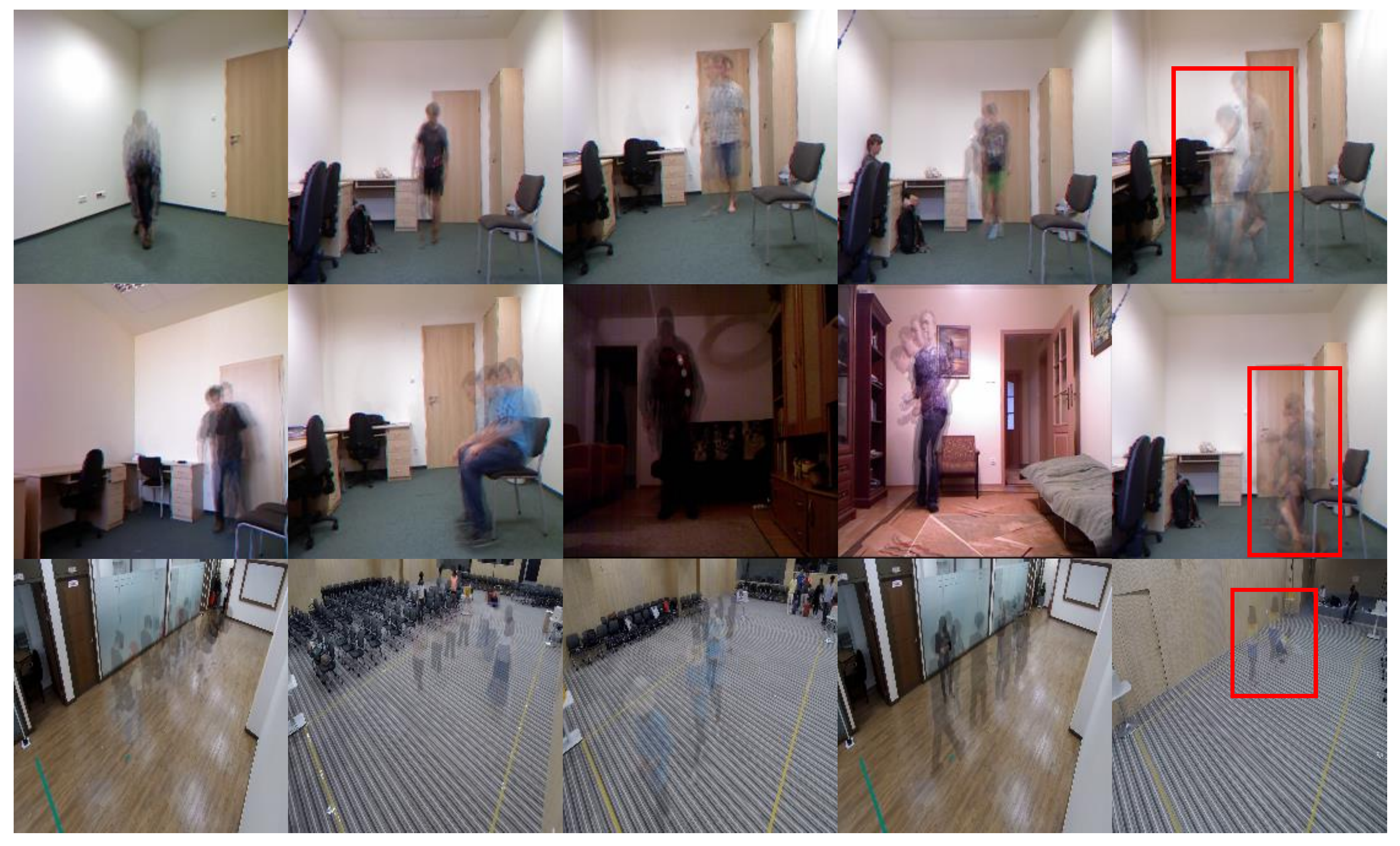}%
    \label{fig:out_nonfalling}%
  }%
  \caption{Examples of the mean frame. Row 1-2 and 3 denote examples of \textit{URFD} and \textit{AIHub} dataset, respectively. Examples of the last column are the failure cases.}
  \label{fig:qualitative}
\end{figure*}

\noindent\textbf{Training and evaluation:}
At the training step, we only use the synthetic data generated by the proposed method.
For the evaluation, we use two different benchmark datasets: \textit{URFD}~\cite{urfd} and \textit{AIHub airport}$^1$.
\textit{URFD} consists of 30 falls and 40 adl (activity of daily living) clips.
All of them are conducted as the test set for the video-level evaluation.
To this end, if each video contains a frame predicted as falling by not less than one frame, the video is considered to be the fall.
Consecutively, if a single frame in the video has never been predicted as falling, it is considered as non-falling.
Experimentally, the interval for the detection was set to 1 second.
\textit{AIHub} consists of 10k clips with three different recording environments.
Among those, we utilize fall and normal videos from \textit{cubox} anomaly scene, which consist of 348 fall clips and 2,109 normal clips with longer duration than \textit{URFD} \ie, 10 seconds or longer with a higher frame rate as 30. 
For this data, the interval for the detection is set to 5 seconds, empirically.
Since the two datasets were collected in different acquisition environments, we utilize two different image blending settings, \ie, scale and transition.

\noindent\textbf{Implementation:}
We utilize OpenCV and albumentation~\cite{albumentation} for the image blending.
Fall detection models are implemented using PyTorch~\cite{pytorch}.
All models are trained about 100 epochs using an AdamW~\cite{adamw} optimizer with a learning rate of 0.001 on a single GPU (RTX 3080).

\begin{table}[ht]
\centering
\caption{Performance of prediction intervals on \textit{URFD}.}
\begin{tabular}{ccc}
\hline
\multicolumn{2}{c}{Interval}                         & \multirow{2}{*}{Accuracy} \\
\multicolumn{1}{c}{Second} & \multicolumn{1}{c}{Frames} &                           \\ \hline
0.4                  & 10                   & 90.0           \\
0.6                  & 15                   & 95.71          \\
0.8                  & 20                   & 95.71          \\
1.0                  & 25                   & \textbf{97.14} \\
1.2                  & 30                   & 91.43          \\
1.4                  & 35                   & 91.43          \\
1.6                  & 40                   & 85.71          \\ \hline
\end{tabular}
\label{tab:buffer} 
\end{table}

\begin{table}[ht]
\centering
\caption{The computational speed of CNNs on \textit{URFD}.}
\begin{tabular}{llll}
\hline
\multirow{2}{*}{Model} & \multicolumn{2}{c}{Frames per second}     & \multirow{2}{*}{Accuracy} \\
 & \multicolumn{1}{c}{Pytorch} & \multicolumn{1}{c}{ONNX} &        \\ \hline
ResNet-18~\cite{resnet}                           & \textbf{707.01}                            & \textbf{1092.80}                         & 94.26             \\
MnasNet~\cite{mnas}                            & 472.22                                     & 1017.88                                 & 92.86             \\
MobileNetV3~\cite{mobilenetv3}                       & 320.34                                     & 722.93                                  & 94.29             \\
EfficientNet-B0~\cite{efficientnet}                  & 162.46            & 204.04                                  & \textbf{97.14}    \\ \hline
\end{tabular}
\label{tab:fps}
\end{table}

\subsection{Results}

\noindent\textbf{Evaluation on \textit{URFD}:}
Table~\ref{tab:acc} shows the fall detection performance of our model compared to existing methods~\cite{pcnn,openpose_fall,maskRcnn,yolov3_deepsort,pipaf,mhi_vgg,bgs_lenet,vgg16_fall,flow_shallow}.
The top five rows are the results of multiple-stage-based approaches, and the other rows are the results of single CNN-based methods.
The proposed method obtains comparable performance to the state-of-the-art methods using multiple modules, in terms of accuracy.
Besides, our method achieves the best performance among single CNN-based methods using EfficientNet-B0~\cite{efficientnet}. 
Table~\ref{tab:buffer} shows the results of the ablation study on the prediction intervals for calculating mean frames. 
Among the intervals, we achieved the best result in one second.
We also compare the computational speed and accuracy of the conducted models as shown in Table~\ref{tab:fps}.
Overall, our method has a slight performance degradation although adopting a model for high speed.

\begin{table}[ht]
\centering
\caption{Fall Detection results on \textit{AIHub}.}
\begin{tabular}{lc}
\hline
\textbf{}          & Frame-level Accuracy          \\ \hline
Sensitivity        & 71.15               \\
Specificity        & 95.76 \\
Precision                 & 73.59  \\
Accuracy                   & 92.25 \\ \hline
\end{tabular}
\label{tab:aihub}
\end{table}

\noindent\textbf{Evaluation on \textit{AIHub}:}
We also evaluate on \textit{AIHub} dataset, as shown in Table~\ref{tab:aihub}.
The results are estimated by frame-level since the falling frames are partial ($<$ 50\%) of whole video clips.
For this dataset, our method achieves an accuracy of 92.25\% for frame-level detection.

\noindent\textbf{Qualitative results:} 
Figure~\ref{fig:qualitative} shows the results on \textit{URFD} and \textit{AIHub}. 
As shown, the falling state is well detected for samples similar to the generated synthetic data.
Besides, ours also correctly predict the samples of multiple persons.
While, ours rarely fails of fall detection in a few samples of a small person or unknown motion, which are not in the training data.

\section{Conclusion}
\label{sec:conclusion}
In this paper, we introduce the new simple fall detection framework.
We propose the simple image blending method to generate human motional and motionless images, representing motion information in a single image.
Then, we train the single CNN with the synthetic training data.
At the inference step, we also represent the human motion in the single frame by estimating the mean image of the selected sequences and using it as input of the detection model. 
Through the extensive evaluation on \textit{URFD} and \textit{AIHub} datasets, we show the proposed method achieves the best performance among baselines. 
While, the handled datasets are containing few humans in one video.
We plan to collect hard samples and expand our research based on localization and object detection tasks to analyze more diverse cases. 

{
\bibliographystyle{IEEEbib}
\bibliography{refers}
}
\end{document}